\documentclass[10pt,twocolumn,letterpaper]{article}

\usepackage{3dv}
\usepackage{times}
\usepackage{epsfig}
\usepackage{graphicx}
\usepackage{amsmath}
\usepackage{amssymb}
\usepackage{multirow}
\usepackage{booktabs}

\DeclareMathOperator*{\argmin}{arg\,min}

\threedvfinalcopy % *** Uncomment this line for the final submission

\def\threedvPaperID{162} % *** Enter the 3DV Paper ID here

\ifthreedvfinal\fi
\begin{document}

%%%%%%%%% TITLE
\title{Depth Completion Using a View-constrained Deep Prior}

\author{Pallabi Ghosh\\
University of Maryland\\
%College Park, MD\\
{\tt\small pallabig@cs.umd.edu}
% For a paper whose authors are all at the same institution,
% omit the following lines up until the closing ``}''.
% Additional authors and addresses can be added with ``\and'',
% just like the second author.
% To save space, use either the email address or home page, not both
\and
Vibhav Vineet\\
Microsoft Research\\
%Redmond, WA\\
{\tt\small Vibhav.Vineet@microsoft.com}
\and
Larry S. Davis\\
University of Maryland\\
%College Park, MD\\
{\tt\small lsd@cs.umd.edu}
\and
Abhinav Shrivastava\\
University of Maryland\\
%College Park, MD\\
{\tt\small abhinav@cs.umd.edu}
\and
Sudipta N. Sinha\\
Microsoft Research\\
%Redmond, WA\\
{\tt\small Sudipta.Sinha@microsoft.com}
\and
Neel Joshi\\
Microsoft Research\\
%Redmond, WA\\
{\tt\small neel@microsoft.com}
}

\maketitle
%\thispagestyle{empty}

%%%%%%%%% ABSTRACT
\begin{abstract}
Recent work has shown that the structure of convolutional neural networks (CNNs) induces a strong prior that  favors natural images.  This prior, known as a deep image prior (DIP), is an effective regularizer in inverse problems such as image denoising and inpainting. We extend the concept of the DIP to depth images. Given color images and noisy and incomplete target depth maps,
we optimize a randomly-initialized CNN model to reconstruct a depth map restored by virtue of using the CNN network structure as a prior combined with a view-constrained photo-consistency loss. This loss is computed using images from a geometrically calibrated camera from nearby viewpoints. We apply this deep depth prior for inpainting and refining incomplete and noisy depth maps within both binocular and multi-view stereo pipelines. Our quantitative and qualitative evaluation shows that our refined depth maps are more accurate and complete, and after fusion, produces dense 3D models of higher quality.
\end{abstract} 
%\vspace{-1cm}
\vspace{-1em}
\section{Introduction}

There are numerous approaches for estimating scene depth, such as using binocular~\cite{middlebury} or multi-view~\cite{furukawa2010towards,goesele2007multi,COLMAPb} stereo, or directly measuring depth with depth cameras, \eg LIDAR, \etc  These approaches suffer from artifacts, such as noise, inaccuracy, and incompleteness, due to various limitations.  As depth estimation is an ill-posed problem, extensive research has been conducted to solve the problem using approximate inference and optimization techniques that employ appropriate priors and regularization~\cite{bleyer2011object,taniai2014graph,taniai2017continuous,woodford2009global}.

%Multi-view stereo~\cite{furukawa2010towards,goesele2007multi,COLMAPb} (MVS) refers to the task of
%computing a detailed 3D model of a stationary scene from multiple calibrated images. Several modern MVS methods
%first compute depth maps from several camera viewpoints using dense stereo matching on multiple sets of %overlapping
%images~\cite{goesele2007multi,COLMAPb,yao2018mvsnet} and then subsequently combines them using a depth-map fusion approach~\cite{fusible}.

Supervised learning methods based on convolutional neural networks (CNNs) have
shown promise in improving depth estimations, both in the binocular~\cite{kendall2017end,mayer2016large,vzbontar2016stereo}
and multi-view~\cite{huang2018deepmvs,ji2017surfacenet,yao2018mvsnet} stereo settings.
However, these supervised methods rely on vast amounts of ground truth data to achieve proper generalization. While unsupervised learning approaches have been explored~\cite{ren2017unsupervised,tonioni2017unsupervised,zhou2017},
their success appears modest compared to supervised methods.

In this work, we propose a new approach for improving depth measurements that is inspired
by the recent work by Ulyanov \etal~\cite{ulyanov2018deep}. They demonstrated that the underlying
structure of a encoder-decoder CNN induces a prior that favors natural images, a property they refer to as a ``deep image prior" (DIP).

%\begin{wrapfigure}{r}{6.0cm}
\begin{figure}[t]
\centering
\includegraphics[width=0.9\linewidth]{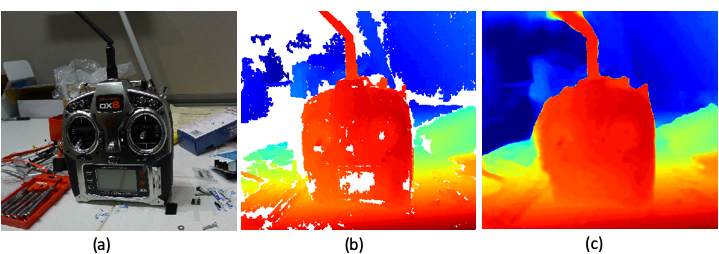}
\caption{(a) Input image from one viewpoint (b) Target depth map computed with SGM using neighboring images\cite{SGM} (c) The refined depth map generated using our deep depth prior (DDP). Depth map (c) has the holes from (b) (shown in white) filled.}
\label{fig:teaser}
\end{figure}
%\vspace{-0.5cm}
%\end{wrapfigure} 

Ulyanov \etal show that the parameters of a randomly initialized encoder-decoder CNN
can be optimized to map a high-dimensional noise vector to a single image.
When the image is corrupted and the optimization is stopped at an appropriate point
before overfitting sets in, the network outputs a remarkably noise-free image.
The DIP has since been used as a regularizer in a number of low-level vision tasks such as image denoising and inpainting~\cite{gandelsman2018double,ulyanov2018deep,williams2019deep}.

In this work, we propose using DIP-based regularization for refining and inpainting
noisy and incomplete depth maps. Our approach can be used to refine depth maps obtained from a wide variety of sources.
%In our case we used an existing efficient stereo matching algorithm~\cite{SGM,mvSGM}.

% (Neel): removed this as this variant it really more of just an ablation of our optimal approach, not a final method we are recommending

%Similar to Ulyanov et al.\cite{ulyanov2018deep}, we also map random noise images to target
%noisy depth maps using randomly initialized encoder-decoder networks- one per depth map.
%In our first variant,
%our network generates a depth map and its parameters are optimized by minimizing a depth reconstruction loss.
%In the other variant, our network outputs an RGB-D image and the reconstruction loss
%is with respect to both color and depth.

Using a network similar to Ulyanov \etal, our approach generates a depth map by combining a depth reconstruction loss with a view-constrained photoconsistency loss.  The latter loss term is computed by warping a color image into neighboring views using the generated depth map and then measuring the photometric discrepancy between the warped image and the original image.

\begin{figure*}[t]
\centering
  \includegraphics[width=0.8\textwidth]{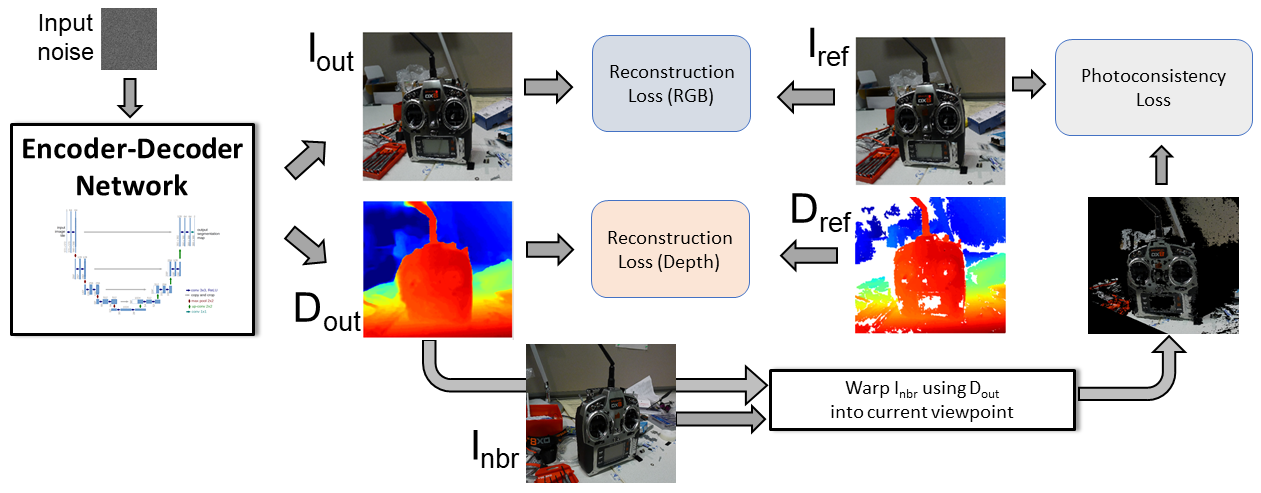}
\caption{Overview of the Deep Depth Prior (DDP): The DDP network is trained using a combination of L1 and SSIM reconstruction loss with respect to a target RGB-D image and a photoconsistency loss with respect to neighboring calibrated images. This network is used to refine a set of noisy depth maps and the refined depth maps are subsequently fused to obtain the final 3D point cloud model. $I^{out}$ and $D^{out}$ are the RGB and depth output of our network. $I^{nbr}$ is the RGB at a neighboring viewpoint. $I^{ref}$ and $D^{ref}$ are the input RGB and depth at the current (reference) viewpoint.}
\label{fig:SystemOverview}
\end{figure*}

In this sense, our technique resembles direct methods proposed decades ago for image registration problems, which all employ some form of initialization and iterative optimization. However, instead of using handcrafted regularizers in the optimization objective,
we use the deep image prior as the regularizer.
While the role of regularization in end-to-end trainable CNN architectures is gaining interest~\cite{knobelreiter2017end,xue2019mvscrf},
our method is quite different, because there is no training and the network parameters are optimized
from scratch on each set of test images. 
Figure~\ref{fig:teaser} shows the inpainting results of applying our technique (DDP) on an input depth map with holes.

To the best of our knowledge, this is the first work to investigate deep image priors for completing depth images. 
%We incorporate the deep depth prior within an approach to refine a depth map given overlapping calibrated color images. In this approach, the regular reconstruction loss of the DIP network is augmented with another loss term which is based on multi-view photoconsistency with respect to the other calibrated color images.
We evaluate our approach using results from modern stereo pipelines and depth cameras and show that the refined depth maps are more accurate and complete, leading to more complete 3D models.

% Since we use deep prior network for 3D reconstruction, there are multiple factors in which our work differs from the original paper. We observed that a combination of mean absolute loss, structural similarity metric and perceptual loss works better than the mean squared error loss used in the original paper. We will discuss this further in the the description and experiments section.

%Our key contribution lies in the deep prior network for 3D reconstruction, and there are multiple factors in which our work differs from~\cite{}. For example, we observe that a combination of mean absolute loss, structural similarity metric, and perceptual loss works better than just the mean squared error loss used in~\cite{}. We will further discuss this in the description and experiments section.

% \begin{figure}
% \centering
%   \includegraphics[width=0.5\textwidth]{Qualitative_results/teaser/Teaser.png}
%   \label{fig:sub1}
% \caption{}
% \end{figure}

% \begin{figure*}
% \centering
%   \includegraphics[width=\textwidth]{Qualitative_results/SystemOverview/System_Overview.pdf}
% \caption{Overview of Deep Reconstruction prior showing the deep encoder decoder and refinement network based on MAE and warping based loss improving depth maps for each view. These depth maps are fused to get a better 3D point cloud reconstruction. }
% \label{fig:SystemOverview}
% \end{figure*}

%\vspace{-6pt}
\vspace{-0.5em}
\section{Related Work}

%In this section, we broadly review existing dense stereo matching and depth refinement methods and then
%discuss the recent work on deep image priors and its applications.

\noindent \textbf{Stereo matching.}
Dense stereo matching is an extensively studied topic and there has been tremendous algorithmic progress
both in the binocular setting~\cite{SGM,knobelreiter2017end,mayer2016large,vzbontar2016stereo,zhou2017} as
well as in the multi-view setting~\cite{fusible,mvSGM,huang2018deepmvs,COLMAPb,yao2018mvsnet}, in conjunction
with advances in benchmarking~\cite{aanaes2016large,geiger2013vision,knapitsch2017tanks,scharstein2014high,schops2017multi}.
Traditionally, the best performing stereo methods were based on approximate MRF inference on pixel grids~\cite{bleyer2011object,taniai2014graph,woodford2009global},
where including suitable smoothness priors was considered quite crucial. However, such methods were usually computationally expensive.
Hirschmuller~\cite{SGM} proposed Semi-Global Matching (SGM), a method that provides a trade-off between accuracy and efficiency
by approximating a 2D MRF optimization problem with several 1D optimization problems. SGM has many recent extensions~\cite{sgmGPU,gehrig2009real,Scharstein2017,seki2017sgm,sinha2014efficient} and also works for multiple images~\cite{mvSGM}.
Region growing methods have also shown promise and implicitly incorporate smoothness priors~\cite{bleyer2011patchmatch,heise2013pm,lu2013patch,COLMAPb}

\noindent \textbf{Deep Stereo.}
In recent years, deep models for stereo have been proposed to compute better matching costs~\cite{chen2015deep,luo2016efficient,vzbontar2016stereo}
or to directly regress disparity or depth~\cite{chang2018pyramid,kendall2017end,mayer2016large,zhou2017} and also for the multi-view setting~\cite{huang2018deepmvs,ji2017surfacenet,yao2018mvsnet}. Earlier on, end-to-end trainable CNN models did not
employ any form of explicit regularization, but recently hybrid CNN-CRF methods have advocated using
appropriate regularization based on conditional random fields (CRFs)~\cite{knobelreiter2017end,xue2019mvscrf}.  In contrast with these works, as we do not perform learning by fitting to training data, our approach is more generalizable as it does not fall prey to the tendency of deep approaches to overfit to their training data.

\noindent \textbf{Depth Map Refinement/Completion.}
The fast bilateral solver~\cite{barron2016fast} is an optimization technique for refining disparity or depth maps. However, the objective
is fully handcrafted. Knoblereiter and Pock recently proposed a refinement scheme where the regularizer in the optimization objective is trained
using ground truth disparity maps~\cite{knobelreiter2019learned}. Their model learns to jointly reason about image color, stereo matching confidence and disparity.
Voynov \etal~\cite{voynov2019perceptual} use a deep prior for depth super-resolution, but they do not have a multiview constraint, as we do, nor do they investigate refinement and hole-filling.
Other recent disparity or depth map refinement techniques utilize trained CNN models~\cite{pang2017cascade}. Similarly depth map completion by Zhang and Funkhouser~\cite{zhang2018deep} use a learning based technique. They do single RGBD image depth completion whereas we use a multi-view photo-consistency loss for training our network. Also we show in out results that one main difference between their work and ours is that our result is not dependent on training data distributions. Depthcomp~\cite{atapour2017depthcomp} also does depth completion and they use the semantic segmentation maps as prior knowledge.

\noindent \textbf{Deep prior for color images.}
Beyond the previously discussed work of Ulyanov \etal~\cite{ulyanov2018deep}, deep image priors have been extended for a number diverse applications -- neural inverse rendering~\cite{taniai2018neural},
mesh reconstruction from 3D points~\cite{williams2019deep}, and layer-based image decomposition~\cite{gandelsman2018double}.
Recently, Cheng \etal~\cite{cheng2019bayesian} pointed out important connections between DIP and Gaussian processes.  Our approach is in a similar vein as these approaches, where we modify the DIP for depth maps by combing the usual reconstruction loss with a second term, the photoconsistency loss
which ensures that when the reference image is warped into a neighboring view using our depth map, the discrepancy
between the warped image and the original image is minimized. 
%\vspace{-6pt}
\section{Method}

Given a RGBD image with $I^\text{in}$ as RGB component and  $D^\text{in}$ as noisy depth component, our goal is to generate denoised and inpainted depth image $D^{*}$.
We leverage recently proposed Deep Image Prior (DIP) ~\cite{ulyanov2018deep} to solve this problem. We first briefly describe the DIP approach. 

\subsection{Deep Image Prior}

The DIP method proposed a deep network based technique for solving low level vision problems such as image denoising, restoration, inpainting, \etc At the core of their method lies the idea that deep networks can serve as a prior for such inverse problems. If $x$ is the input image, $n$ is the input noise  and $x_{o}$ is the denoised output of the network $f_{\theta}$, then the optimization problem of the DIP method takes the following form:
\begin{equation} 
\theta^*=\argmin_{\theta} L(f_{\theta}(n); x), \; \; \; x_{o}^*=f_{\theta^{*}}(n). 
\end{equation}

The task of finding the optimal neural network parameters $\theta^*$ and the optimal denoised image $x_o^*$ is solved using the standard backpropagation approach.

\begin{figure}[t]
\centering
\includegraphics[width=0.9\linewidth]{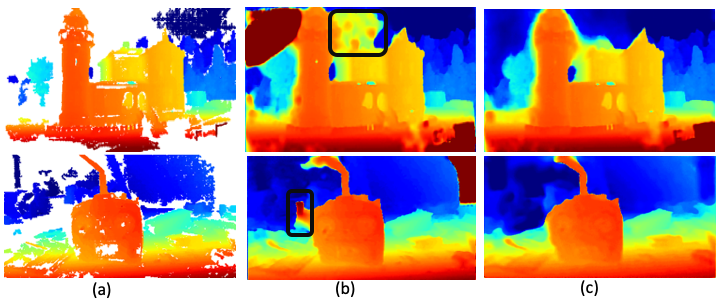}
\caption{(a) Input depth map with holes (b) DDP on just depth maps and (c) DDP on RGBD images. In the black box regions in (b), DDP is filling up the holes in the sky or background based on the depth from the house or radio because it has no edge information. RGBD input provides this edge information in (c).}
\label{fig:DvsRGBD}
%\vspace{-0.75cm}
\end{figure}

A simple approach to address depth denoising and inpainting would be to use a DIP like encoder-decoder architecture to improve the depth images. Here depth images would replace RGB as inputs in the original DIP framework. However this fails to fill the holes with correct depth values. Some of the results are shown in Figure~\ref{fig:DvsRGBD}. More quantitative results are provided in Table \ref{table:ablation}.
We hypothesize three reasons for this failure. First, holes near object boundary can cause incorrect depth filling. 
Second, depth images have more diverse values than RGB images that leads to large quantization errors. %Correctly predicting values for such large quantized space is a challenging task for the DIP network. 
Finally absolute error for far objects may dominate the DIP optimization over important nearby objects.

\subsection{Deep Depth Prior}

In this work, we propose the Deep Depth Prior (DDP) and introduce three losses to solve the issues discussed above.
Our approach is built on top of the inpainting task in Ulyanov \etal~\cite{ulyanov2018deep} where we create a mask for the holes in the depth map and calculate the loss over the non-masked regions. Figure~\ref{fig:SystemOverview} gives an overview of our system.

To solve the issue of absolute error for far objects dominating the DIP optimizer, inverted depth or disparity images are used.
We also add a constant value to depth image that reduces the ratio between the maximum and minimum depth values. 
Further, masking of all far away objects beyond a certain depth is performed by clipping to a predefined maximum depth value. We also clip depth to a minimum value so that the maximum disparity value does not go to infinity. We provide these values and values of the loss hyper-parameters in the supplementary. %\neel{What are these values? List them somewhere in the paper}

%Let $f_{\theta}$ denote our encoder-decoder generator network that outputs the depth $d^\text{out}$.
Let $D^\text{out}$ be the desired depth output from our network and let $f_{\theta}$ denote the generator network.
Input to the network is noise $n^\text{in}$ and the input depth map is inverted to get $Z^\text{in}$ as the noisy disparity map. Let us represent the output from the network as $Z^\text{out}$ where $Z^\text{out} = f_{\theta}(n^\text{in}; Z^\text{in})$. On convergence, optimal $D^{*}$ is obtained by inverting $Z^{*}$.

We use three different losses to optimize our network. The total loss is defined as follows.
\begin{equation}
%\vspace{-2mm}
L^\text{total}= \gamma_{1}L^\text{disp}+\gamma_{2}L^\text{RGB}+(1-\gamma_{1}-\gamma_{2})L^\text{warp}.
\label{eq:tl}
\end{equation}

\vspace{-1em}
%\vspace{-3mm}
\paragraph{\bf Disparity-based loss ($L^\text{disp}$).}

The simplest technique to obtain $Z^{*}$ is to optimize only on disparity. The disparity based loss $L^\text{disp}$ is a weighted combination of Mean Absolute Error (MAE) or $L^{1}$ loss and Structural Similarity metric (SSIM) or $L^\text{SSIM}$ loss \cite{ssim2004tpami},  and takes the following form:
\begin{equation}
\begin{aligned}
L^\text{disp} = \lambda_{z}L^{1}(Z^\text{in},Z^\text{out})+
(1-\lambda_{z})L^\text{SSIM}(Z^\text{in},Z^\text{out}).
\end{aligned}
\end{equation}

We use $L^{1}$ loss instead of Mean Squared Error($L^{2}$) loss to remove the effect of very high valued noise having a major effect on the optimization. It takes the form as  $L^{1}(Z^\text{in},Z^\text{out}) = |Z^\text{in}-Z^\text{out}|$.
The structural similarity $L^\text{SSIM}$ loss measures similarity between the input $Z^\text{in}$ and reconstructed disparity map $Z^\text{out}$. Here similarity is defined at the block level where each block size is 11x11. %\Vibhav{provide accurate value here}. 
It provides consistency at the region level.
%helps to put more weight to the loss in regions of structural differences in the image like edges instead of equal weight to all locations.
The loss ($L^\text{SSIM}$) takes the following form $L^\text{SSIM} = 1 - SSIM(Z^\text{in},Z^\text{out}).$
%
%\begin{equation}
%L^\text{SSIM} = 1 - SSIM(z^\text{in},z^\text{out}).
%\end{equation}
Details about $SSIM$ are provided in the supplementary.
%\vspace{-3mm}

\vspace{-1em}
\paragraph{RGB-based loss ($L^\text{RGB}$).}

We observed that under certain situations the disparity based loss leads to blurred edges in the final generated depth map. This happens when there is a hole in the depth map near an object boundary. The generator network produces a depth map that fuses the depths of different objects appearing around the hole. 

%This happens especially when there is an unfilled region in the depth map beside a well detected region, and the unfilled region belongs to a separate surface at a different depth.

%For example in Figure~\ref{fig:DvsRGBD}, the house has well detected points and depth map, but the sky does not have definite feature match across images due to homogeneity. 
%For example in Figure~\ref{fig:DvsRGBD}, the sky has holes in the depth map because of homogeneity.

For example, consider regions belonging to sky in the top row of Figure~\ref{fig:DvsRGBD}. Due to the homogeneous nature of the sky pixels, standard disparity/depth estimation methods fail to produce any valid values for such regions. However, pixels corresponding to house region have depth values. 
%Hence ambiguous regions in the sky have no depth. 
When the image is passed to a DIP generator, the edge between the house and the sky gets blurred because the network is trying to fill up the space without any additional knowledge, e.g., boundary information. It just bases its decision on depth of neighboring space %\neel{homogeneity of space?} 
to fill up the incomplete regions as seen in the top row of Figure~\ref{fig:DvsRGBD}(b) in the black box region.

To solve this problem, we also pass the color RGB image along with the disparity image. The encoder-decoder based DDP architecture is now trained on the 4 channel RGBD image. The network weights are updated not only on the masked disparity map but also on the full RGB image. This helps the network to leverage edge and texture information for the object boundary in the RGB image to fill the holes in the disparity (and so in depth) image. 
This important edge information provided to the network helps to generate  crisp depth images as seen in the Figure~\ref{fig:DvsRGBD}(c). 

Let $I^\text{out}$ be the output corresponding to input data $I^\text{in}$ using the noise $n^\text{in}$ and generator model $f_{\theta}$. It takes the form as $I^\text{out} = f_{\theta}(n^\text{in}; I^\text{in})$.
The Loss $L^\text{RGB}$ is also a weighted combination of $L^1$ and $SSIM$ losses, and is defined as:
\begin{equation}
\begin{aligned}
L^\text{RGB} =
\lambda_{I}L^{1}(I^\text{in},I^\text{out})
+ (1-\lambda_{I})L^\text{SSIM}(I^\text{in},I^\text{out}).
\end{aligned}
\end{equation}

\begin{figure}[t]
\centering
  \includegraphics[width=0.9\linewidth]{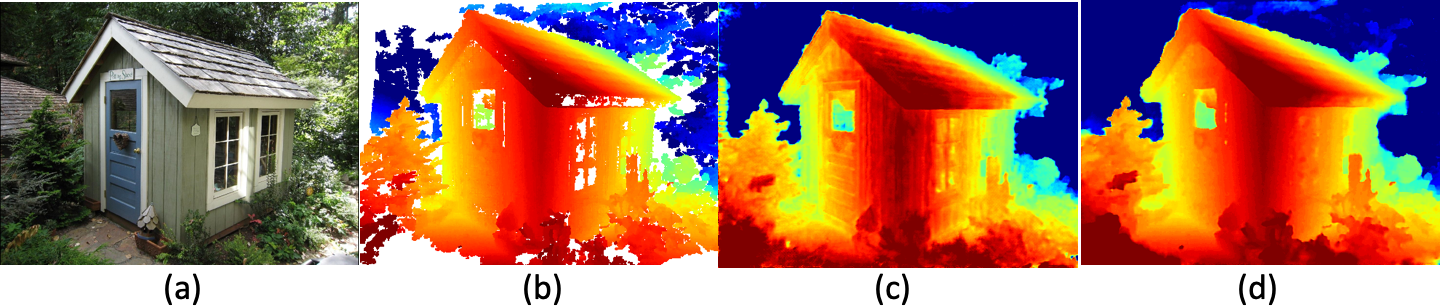}
\caption{(a) RGB image (b) Input disparity map (c) Disparity output from DDP trained with equal weight for RGB and depth loss. The RGB artifacts are evident in (c) through the vertical and horizontal lines representing the wooden planks in the wall (d) Disparity output from the DDP trained with lower weight for the RGB loss compared to depth loss. The artifacts disappear in (d).}
\label{fig:HighRGBWeight}
\end{figure}

The RGB based loss helps to resolve the issue of blurring observed around edges near object boundaries. However, putting equal weights to disparity ($L^\text{disp}$) and RGB  ($L^\text{RGB}$) components of the total loss leads to artifacts appearing in the disparity image and so in depth output images as well. In particular, these artifacts are due to textures and edges from an object's appearance that are unrelated to the depth boundaries. For example in Figure~\ref{fig:HighRGBWeight} the wall of the house is one surface and should have smooth depth maps. 
However, the DDP network trained on RGBD data generates vertical textures in the depth images that appear due to the vertical wooden planks in the RGB image.
%Next we describe warping based consistency loss that resolves this issue..
%
\vspace{-1em}
%\vspace{-3mm}
\paragraph{\bf Warping-based loss ($L^\text{warp}$).}
Lastly, we include a warping loss.
Before defining the warping loss, let us first define the warping function $T_\text{nbr}^\text{ref}$. Given the camera poses $C_\text{ref}$ and $C_\text{nbr}$ of the reference and neighboring view, the function $T_\text{nbr}^\text{ref}$ warps neighboring view to reference view.

We are trying to generate denoised output $Z^\text{out}$ for the reference view. We first find top $N$ neighboring views of the reference view using the method used in MVSNet~\cite{yao2018mvsnet} for the multi-view pipeline. 
%
%We use the same technique as MVSNet to compute the top $N$ views to use for warping into the current view {j}. 
Let $nbr$ denote one of these N views and let $W_\text{nbr}^\text{ref}$ be the warped image from neighboring view to reference view. Let $D^\text{out}_\text{ref}$ be the predicted depth (inverted $Z^\text{out}_\text{ref}$) for the reference view and $I^\text{in}_\text{nbr}$ is the input RGB for the neighboring view, then the warped image is $W_\text{nbr}^\text{ref} = T_\text{nbr}^\text{ref}(D^\text{out}_\text{ref},I^\text{in}_\text{nbr}; C_\text{nbr}, C_\text{ref})$.
%
%\begin{equation}
%\begin{aligned}
%W_{i}^{j} = T_{i}^{j}(d^\text{out}_{j},I^\text{in}_{i})
%\end{aligned}
%\end{equation}
%
Further, we use bilinear interpolation while warping. %\neel{Don't we use bilinear interpolation for all the warping, not just to remove holes?  Or is there a separate hole removal step?}

Now given $I^\text{in}_\text{ref}$, the input RGB for the reference view, we can compute warping loss as:
\begin{equation}
\begin{aligned}
L^\text{warp}_\text{nbr-ref} =
\lambda_{w}L^\text{SSIM}(I^\text{in}_\text{ref},W_\text{nbr}^\text{ref})
+ (1-\lambda_{w})L^{1}(I^\text{in}_\text{ref},W_\text{nbr}^\text{ref}).
\end{aligned}
\end{equation}

When there are multiple neighboring views, the loss is averaged over them as $L^\text{warp}_\text{ref} = \frac{1}{N}\sum_\text{nbr=1}^{N}L^\text{warp}_\text{nbr-ref}.$

%The warping loss not only resolves the issue of artifacts appearing around edges within objects, it helps to improve the accuracy of the disparity (and depth) values in other regions as well. 
The warping loss helps to improve the accuracy of the disparity (and depth) values and the importance of each loss term is explored further in section \ref{sec:exp}. 

\vspace{-1em}
\paragraph{Optimization.} All three losses that we use are differentiable with respect to the network parameters and so the network is optimized using standard backpropagation. %\neel{Say here what is the network structure, optimizer, learning rate, number of iterations/epochs.  Or maybe put in supplemental and refer to that here and/or refer to the structure and parameters used in the DIP paper.}

\section{Experiments}
\label{sec:exp}

We demonstrate the effectiveness of our proposed approach on two different tasks - 1) depth completion and 2) depth refinement. 
We also show the generalization ability of our technique on new datasets with unseen statistical distributions. 
%\paragraph{\bf Implementation.} 
We applied the deep depth prior on multi-view and stereo pipelines and evaluate results on five different datasets: 1) Tanks and Temples (TnT) \cite{knapitsch2017tanks}, 2) KITTI stereo benchmark \cite{geiger2013vision,Menze2015CVPR}, 3) Our own collected videos, 4) NYU depth V2 \cite{eccv_SilbermanHKF12} and 5) Middlebury Dataset \cite{hirschmuller2007evaluation,scharstein2007learning}. %a sequence of depth images of a static scene.
In all our experiments, the base network is primarily an encoder-decoder (UNet) architecture \cite{ronneberger2015u}. 
The UNet encoder consists of 5 convolution blocks each consisting of 32, 64, 128, 256, and 512 channels. Each convolution operation uses 3x3 kernels. 
We also conducted experiments using skip networks \cite{ulyanov2018deep}.  
The input noise to the network is of size $m \times n \times 16, $ where $m$ and $n$ are the dimensions of the input depth images. %960x512x16 %\Vibhav{@Pallabi, update this number}.
Training is performed using the Adam optimizer. Details about learning rate and scheduler are provided in the supplementary.  %Hyper-parameters of the loss function are set through searching on a small subset (10 images randomly chosen) from each of Ignatius and Barn sequence in TnT. %the TnT dataset. %\neel{how small?  what was used/picked and why?} %\Vibhav{Confirm this}.
%
%
%We also show the generalization power of our technique to new datasets with unseen statistical distributions. 
%

\subsection{Tanks and Temples}
\begin{table}[t]
\centering
% \begin{minipage}{.47\textwidth}
  \caption{We compare the f-score of DDP on datasets of different sizes. As the number of images become smaller, the holes increase and the most relative performance gain is at 22 images. We also compare to \cite{zhang2018deep} applied to an data that is out of distribution and show we do better. Higher is better.}
  %\resizebox{\linewidth}{!}{
    \begin{tabular}{@{}lcccc@{}}
    \toprule
    %\begin{tabular}{|l|c|c|c|c|}
    %\hline
    Dataset & SGM & SGM+ & SGM+ \\
    & & \cite{zhang2018deep} &  DDP (Ours) \\
    \hline
    % Ignatius( 85 images) & 44.40 & 44.35 & 44.70  \\
    % Ignatius( 43 images) & 43.36 & 43.35 & 43.52 \\
    % Ignatius( 22 images) & 31.66 & 31.60 & 32.17 \\
    % Ignatius( 11 images) & 30.70 & 30.66 & 31.25 \\
    Ignatius( 87 images) & 45.3 & 45.2 & 45.6  \\
    Ignatius( 44 images) & 44.0 & 44.0 & 44.2 \\
    Ignatius( 22 images) & 30.7 & 30.7 & 31.1 \\
    %Ignatius( 11 images) & 23.7 & 23.7 & 23.9 \\
    %\hline
    \bottomrule
    \end{tabular}
    \label{table:skip_size}
    \end{table}

\begin{table}[t]
% \end{minipage}%
% \hspace{1em}
% \begin{minipage}{.47\textwidth}
    \centering
    \caption{We show results of comparing the DDP output using disparity, RGBD, and warping loss and using UNet/SkipNet. P: precision, R: recall, F: f-Score. Higher is better.}
    \begin{tabular}{@{}lcccc@{}}
    \toprule
    Network + Loss & P & R & \textcolor{blue}{F} \\
    \midrule
    % UNet+D & 32.33 & 35.50 & \textcolor{blue}{33.84} \\
    % UNet+RGBD & 38.50 & 38.73 & \textcolor{blue}{38.62} \\
    % UNet+RGBD+warp & 42.83 & 49.75 & \textcolor{blue}{\bf{46.03}} \\
    % SkipNet+RGBD+warp & 42.81 & 49.77 & \textcolor{blue}{46.03} \\
    UNet + D & 34.3 & 37.2 & \textcolor{blue}{35.7} \\
    UNet + RGBD & 39.2 & 49.0 & \textcolor{blue}{43.6} \\
    UNet + RGBD + warp & 40.0 & 50.6 & \textcolor{blue}{\bf{44.7}} \\
    SkipNet + RGBD + warp & 40.2 & 50.3 & \textcolor{blue}{44.7} \\
    %{\bf A4} DDP+UNet+Refinement & 41.84 & 48.95 & 45.12 \\
    \bottomrule
    \end{tabular}
    \vspace{-0.1in}
    \label{table:ablation}
% \end{minipage}
\end{table}

% \VIBHAV{In tables, we should find a way to show that higher number is better - either as text or as arrow.}

We evaluate the effectiveness of the presented approach on the multi-view reconstruction task through qualitative and quantitative improvement on seven sequences from the Tanks and Temples dataset (TnT dataset) \cite{knapitsch2017tanks}. These sequences include Ignatius, Caterpillar, Truck, Meetingroom, Barn, Courthouse, and Church. To reconstruct the final 3D point cloud the depth images are fused using the approach proposed by Galliani \etal~\cite{galliani2015massively} (Fusibile). Fusibile has hyperparameters that determine the precision and recall values for the resultant point cloud. These parameters include the disparity threshold and the number of consistent views. More details about Fusibile and loss hyper-parameters are provided in the supplementary material.

\vspace{-1em}
\paragraph{\bf Quantitative results.}
Our primary comparison is with the popular semi-global matching (SGM) method \cite{SGM} for depth image estimation. SGM is an optimization based method that does not need any training data. We also compare with a state-of-the-art learning based method: MVSNet \cite{yao2018mvsnet}. Further, it should be noted that our approach is agnostic to the depth estimation method, i.e., it can be used to improve depth maps from any source. 

We compare our reconstructed point clouds to the ground truth point clouds for all 7 sequences in the TnT dataset~\cite{knapitsch2017tanks}, using the benchmarking code included with the dataset, which returns the precision ($\textrm{P}=\frac{\textrm{TP}}{\textrm{TP}+\textrm{FP}}$), recall ($\textrm{R}=\frac{\textrm{TP}}{\textrm{TP}+\textrm{FN}}$) and f-score ($\textrm{F}=2.\frac{\textrm{P.R}}{\textrm{P}+\textrm{R}}$) values for each scene given the reconstructed point cloud model and a file containing estimated camera poses used for that reconstruction. Here, $TP$, $FP$, and $FN$ are true positives, false positives, and false negatives respectively. For each of the sequences we report values at the same points of the precision-recall curve as specified by the TnT dataset.%they also fix the point where precision and recall are reported given the precision-recall curve.
\vspace{-1em}
\paragraph{\bf Ablation studies.}
\label{AS}

\begin{table}
\centering
% \small
\caption{Quantitative results comparing 7 sequences for SGM based depths and applying DDP on SGM depths. We combine DDP with SGM by replacing the depth values in the holes of SGM depth with DDP depth. Here the datasets are I: Ignatius, B: Barn, T: Truck, C1: Caterpillar, MR: Meetingroom, CH: Courthouse, and C2: Church. N: number of images in a sequence, C: number of consistent views while constructing point cloud, D: disparity threshold, P: precision, R: recall, F: f-score. Higher is better.}
\renewcommand{\arraystretch}{1.1}
\resizebox{\linewidth}{!}{
\begin{tabular}{@{}llcccccccc@{}}
\toprule
&  &  &  & \multicolumn{3}{c}{SGM} & \multicolumn{3}{c@{}}{SGM+DDP(Ours)} \\
\cmidrule(lr){5-7}
\cmidrule(lr){8-10}
Data& N & C & D & P & R & \textcolor{blue}{F} & P & R & \textcolor{blue}{F} \\
\midrule
% Ignatius & 68 & 4 & 1.0 & 43.97 & 48.68 & \textcolor{blue}{46.20} & 43.54 & 50.01 & \textcolor{blue}{\bf{46.55}} \\
% Ignatius & 17 & 1 & 2.0 & 30.17 & 32.76 & \textcolor{blue}{31.41} & 29.32 & 34.33 & \textcolor{blue}{\bf{31.63}} \\
I & 87 & 5 & 1.0 & 41.7 & 49.5 & \textcolor{blue}{45.3} & 41.2 & 51.1 & \textcolor{blue}{\bf{45.6}} \\
I & 22 & 2 & 2.0 & 32.7 & 29.0 & \textcolor{blue}{30.7} & 32.2 & 30.1 & \textcolor{blue}{\bf{31.1}} \\
B & 180 & 2 & 0.5 & 23.3 & 27.8  & \textcolor{blue}{25.4} & 22.8 & 29.3 & \textcolor{blue}{\bf{25.6}} \\
B & 45 & 1 & 4.0 & 19.1 & 21.4 & \textcolor{blue}{20.2} & 18.1 & 22.8 & \textcolor{blue}{\bf{20.2}} \\
T & 99 & 4 &  1.0 & 35.8 & 38.8 & \textcolor{blue}{37.2} & 34.5 & 41.4 & \textcolor{blue}{\bf{37.6}} \\
T & 25 & 1 &  2.0 & 29.4 & 33.8 & \textcolor{blue}{31.5} & 27.7 & 36.7 & \textcolor{blue}{\bf{31.6}} \\
C1 & 156 & 4 & 1.0 & 24.9 & 41.5 & \textcolor{blue}{\bf{31.1}} & 24.0 & 42.9 & \textcolor{blue}{30.8}  \\
C1 & 39 & 1 & 2.0 & 17.3 & 36.5 & \textcolor{blue}{\bf{23.5}} & 16.2 & 37.9 & \textcolor{blue}{22.7}  \\
MR & 152 & 4 & 1.0 & 27.5 & 13.4 & \textcolor{blue}{18.1} & 25.2 & 15.2 & \textcolor{blue}{\bf{19.0}}\\
MR & 38 & 1 & 4.0 & 17.8 & 9.0 & \textcolor{blue}{12.0} & 15.7 & 10.7 & \textcolor{blue}{\bf{12.7}}\\
CH & 110 & 2 & 1.0 & 1.8 & 0.8 & \textcolor{blue}{1.1} & 3.2 & 1.2 & \textcolor{blue}{\bf{1.7}} \\
C2 & 86 & 4 & 1 & 8.9 & 8.5 & \textcolor{blue}{8.7} & 8.9 & 8.6 & \textcolor{blue}{\bf{8.8}} \\
\bottomrule
\end{tabular}
}
\vspace{-0.1in}
\label{Table:SGM}
\end{table}

% \begin{table}[t]
% \centering
% \begin{tabular}{|l|c|c|c|c|c|c|c|}
% \hline
% Dataset & \multicolumn{3}{c|} {MVSNet} & \multicolumn{3}{c|} {MVSNet+DDIP}\\
% & P & R & \textcolor{blue}{F} & P & R & \textcolor{blue}{F} \\
% \hline\hline
% Ignatius & 45.9 & 52.2 & \textcolor{blue}{48.8} & 45.2 & 54.9 &
% \textcolor{blue}{\bf{49.6}}  \\
% (126 imgs) & & & & & & \\
%  %& & & & & & \\
% Ignatius & 40.3 & 46.8 & \textcolor{blue}{43.3} & 38.9 & 50.0 & \textcolor{blue}{\bf{43.9}}  \\
% (32 imgs) & & & & & & \\
%  %& & & & & & \\
% \hline
% \end{tabular}
% \caption{We compare the reconstruction performance using depth map generated by MVSNet and depth map generated by applying DDP on MVSNet. P:precision, R:recall, F:f-score. Higher is better.}
% \label{table:MVSnet_DDIP}
% \end{table}
We conduct a series of experiments on the Ignatius dataset to study impact of each parameter and component of the proposed method. We first study the effect of number of depth images on the final reconstruction. For this purpose, we skip a constant number of images in the dataset. %vary a parameter, {\em jump size}, that controls the number of images that are used for final reconstruction. 
For example, a skip size of 2 means that only every second image is used for reconstruction giving us 87 images in total for Ignatius. Such reduction in the data size increases the number of holes on the SGM-based reconstruction method. The same reduced dataset is used for the baseline and our approach. 
One of the goals of this work is for our accurate hole-filling to allow for fewer images to be captured and used for reconstruction. We also conducted experiments with skip values of 4 and 8 giving 44 and 22 images. 
Table \ref{table:skip_size} provides details about the impact of this skip size on relative improvement. For SGM+DDP, we keep SGM values where there are no holes, and replace the holes with DDP output values in those regions. It can be observed that at skip sizes of 2, 4 and 8, we see an improvement of 0.3, 0.2 and 0.4 percentage point of relative improvement in f-scores over the baseline respectively. %\Vibhav{@Pallabi, can you check if the numbers in the table for jump size are final. Then accordingly update the previous number in the text.} 
It suggests that at higher skip sizes, our approach provides necessary prior information to fill holes. Please note that we have not included experiments with skip size of one. Running Fusibile on all of the Ignatius images produces dense reconstruction without holes. Running DDP on this dense reconstruction has no effect, and so we have not included results with skip size of one in the table. %One of the goals of this work is for our accurate hole-filling to allow for fewer images to be captured and used for reconstruction.
Table~\ref{table:skip_size} also contains results in comparison to Zhang and Funkhouser~\cite{zhang2018deep} which we will discuss later.

Next we show the advantage of using the RGBD and warping losses over disparity loss only. Running the optimization for too many epochs on the depth only DDP can return the holes in the result, thus we run the depth only based networks for 6000 epochs instead of 16000. The results are in Table~\ref{table:ablation}, showing a $7.9\%$ improvement in f-measure by using RGBD. Table~\ref{table:ablation} also shows the benefit of photo-consistency based warping loss. We observe a relative improvement of $1.1\%$ in f-measure after incorporation of the warping loss. 
Finally, we also conducted experiments with Skip-Net \cite{ulyanov2018deep} to show the impact of using a network different from UNet.  Table~\ref{table:ablation} shows they give comparable results, and we chose to use UNet for our other experiments since it is a more commonly used network. %\neel{Why isn't the 85 image result in Table 1 the same as Unet+RGBDwarp in table 2?  Is there some difference? I'd expect there is some baseline best result for ignatius that we then degrade with each ablation.  Does table 1 include the swapping in of the original SGM results for where there weren't holes?  This is discussed below, but mention it for the table 1 result if it is used..}

\begin{table}[t]
\centering
\caption{We compare the reconstruction performance using depth maps generated by MVSNet and by applying DDP on MVSNet. I: Ignatius, P: precision, R: recall, F: f-score. Higher is better.}
\resizebox{\linewidth}{!}{
\begin{tabular}{@{}lccccccc@{}}
\toprule
Dataset & \multicolumn{3}{c@{}} {MVSNet} & \multicolumn{3}{c@{}} {MVSNet + DDIP}\\
\cmidrule(lr){2-4}
\cmidrule(lr){5-7}
& P & R & \textcolor{blue}{F} & P & R & \textcolor{blue}{F} \\
\midrule
I (126 images) & 45.9 & 52.2 & \textcolor{blue}{48.8} & 45.2 & 54.9 &
\textcolor{blue}{\bf{49.6}}  \\
 %& & & & & & \\
I (32 images) & 40.3 & 46.8 & \textcolor{blue}{43.3} & 38.9 & 50.0 & \textcolor{blue}{\bf{43.8}}  \\
 %& & & & & & \\
\bottomrule
\end{tabular}}
\vspace{-0.1in}
\label{table:MVSnet_DDIP}
\end{table}

\vspace{-1em}
\paragraph{\bf Comparison to prior work.}
\label{comp}
Quantitative comparison with the SGM baseline on the 7 TnT dataset is shown in the Table \ref{Table:SGM}. The SGM+DDP (Ours) column shows the results of using DDP to improve depth maps from SGM. We combine DDP depth maps with SGM depth maps, keeping the SGM values everywhere where there are no holes and replace the holes with DDP depth values. The table includes the precision, recall and f-scores on each dataset with sub-sampled number of images using the technique described in the Ablation Studies sub-section. We observe improvement in both recall and f-score values of the presented approach over the SGM-method. It suggests the method helps in hole filing. The f-score improves in 6 of the 7 sequences. Overall SGM+DDP(Ours) helps to improve recall by $1.5$ percentage points and f-score by $0.2$ percentage points. In particular, we see a significant improvement of $1.8$ in recall and $0.8$ in f-score on the MeetingRoom sequence.

\begin{figure}
\centering
  \includegraphics[width=0.9\linewidth]{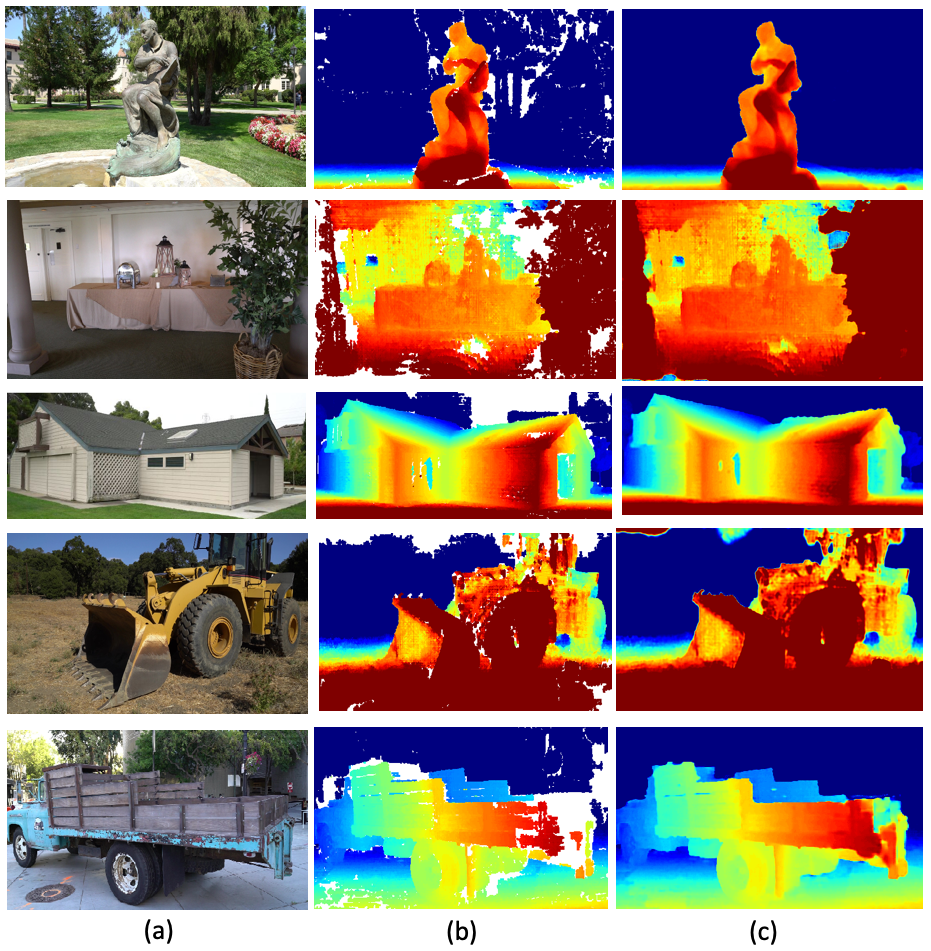}
  \caption{(a) Input RGB and (b) Input depth images and (c) Predicted depth at 16000 epochs for Ignatius, Meeting Room, Barn, Caterpillar and Truck. (Best viewed in digital. Please zoom in.)}
\label{fig:depthmaps}
\vspace{-0.1in}
%\vspace{-0.5cm}
\end{figure}

% \begin{wraptable}{R}{7.5cm}
% \caption{Results comparing our output applied on SGM depth to MVSNet depth images. P:precision, R:recall and F:f-score. We have applied our results on 5 TnT training datasets, Ignatius, Barn, Truck, Caterpillar and MR: Meetingroom sequence. Higher is better.}\label{Table:MVSNet}
% \begin{tabular}{|l|c|c|c|c|c|c|c|}
% \hline
% Dataset & \multicolumn{3}{c|} {SGM+DDP(Ours)} & \multicolumn{3}{c|} {MVSNet}\\
% & P & R & \textcolor{blue}{F} & P & R & \textcolor{blue}{F} \\
% \hline\hline
% Ignatius & 43.5 & 50.0 & \textcolor{blue}{46.6} & 37.2 & 63.0 & \textcolor{blue}{46.8} \\
% Barn & 18.1 & 22.8 & \textcolor{blue}{20.2} & 19.1 & 18.8 & \textcolor{blue}{18.9} \\
% Truck & 34.5 & 41.4 & \textcolor{blue}{37.6} & 29.1 & 51.1 & \textcolor{blue}{37.1} \\
% Caterpillar & 16.2 & 37.9 & \textcolor{blue}{22.7} & 25.2 & 32.9 & \textcolor{blue}{28.5} \\
% MR & 25.3 & 15.2 & \textcolor{blue}{19.0} & 18.1 & 11.9 & \textcolor{blue}{14.4} \\
% \hline
% \end{tabular}
% \end{wraptable}
% Table~\ref{Table:MVSNet} compares precision, recall and f-scores of SGM+DDP (Ours) method with that of the learning based MVSNet baseline. It is based on applying DDP on SGM depth and not MVSNet depth which means our technique is not learning dependent whereas MVSNet is. Our approach achieves an improvement of $0.1$ percentage points for f-score over the MVSNet method.  
We also test using the learning-based MVSNet method as an input to our method. While the results from MVSNet do not contain any holes, as they predict a value at every pixel, just as we do, they do have areas where the predictions have low confidence. We use their output probability map which shows confidence of depth prediction and remove depths at places with confidence below a certain threshold (0.1) to see if we can fill those areas more accurately than MVSNet did. We then fill up these holes using DDP and compare the results with the original. These results for Ignatius are in Table~\ref{table:MVSnet_DDIP}. We see $0.7\%$ improvement in F-score and $3.0\%$ in recall.

Finally we compare to Zhang and Funkhouser~\cite{zhang2018deep} in Table~\ref{table:skip_size}. We use the trained networks provided by Zhang and Funkhouser using SUNCG-RGBD~\cite{song2015sun} and ScanNet~\cite{dai2017scannet} datasets for inpainting on Ignatius dataset. As we can see here this method does not work well because it is a learning based system. Ignatius is an out-of-distribution test data and the model would require finetuning. This emphasizes the usefulness of our network being optimized at test time and it being independent of training data statistics.
\vspace{-1em}
\paragraph{\bf Qualitative results.}

Next we provide visual results on the TnT dataset to highlight the impact of our approach in achieving high quality reconstruction. 
In Figure~\ref{fig:depthmaps}, we show output disparity images at 16000 (column c) epochs of our proposed approach on 5 TnT sequences. Note that the holes in the input disparity images (marked as white in column b) are filled in the output images (c). 
% Further, qualitative improvement can be observed across epochs on comparing visual outputs. For example, disparity of the background wall in second row image is more consistent at 16000 epochs than at 500 epochs. 

\subsection{KITTI}

\begin{table}
%\vspace{-1cm}
\centering
\caption{Results on KITTI Dataset using D1 error. Lower is better.}
\begin{tabular}{@{}lcccc@{}}
        \toprule
         Method & D1-bg & D1-fg & D1-all \\
         \midrule
         ELAS (params1) & 7.5 & 21.1 & 9.8 \\
         ELAS+DDP (params1) & 7.5 & 21.1 & 9.7 \\
          %& & & \\
         \midrule
         ELAS (params2) & 20.6 & 28.7 & 22.0 \\
         ELAS+DDP (params2)& 12.3 & 20.2 & 13.6 \\
         % & & & \\
         \midrule
         HD3 & 1.78 & 3.64 & 2.09 \\
         HD3+DDP & 1.75 & 3.55 & 2.05 \\
         HD3 trained on~\cite{mayer2016large}& 78.74 & 79.16 & 78.81 \\
         %on Things & & & \\
         \bottomrule
    \end{tabular}
\label{tab:KITTI}
\vspace{-0.1in}
%\vspace{-0.5cm}
\end{table}

Next we show the performance of the proposed approach on the KITTI stereo benchmark (2015)~\cite{Menze2015CVPR} for 3 different input disparity maps in Table\ref{tab:KITTI}.
The evaluation metrics used is D1 that measures the error percentage when the predicted value differs from the groundtruth value by 3 pixels or 5\% or more. This error is separately measured for background, foreground and all regions together. 
%Three input depths used are listed below.
3 approaches to generate input depths for DDP are as follows:
\begin{enumerate}

\item We use first set of parameters for ELAS~\cite{geiger2010efficient} to generate disparity maps without holes and DDP for refinement and completion. We observe a small performance improvement ($\sim$0.1\% reduction in D1-all error).

\item We use a second set of parameters for ELAS to generate depth images with holes. In this case, DDP applied over ELAS depth helps to inpaint and complete the ELAS depth with a 8.4\% reduction in D1-all error.

\item We take the output of a state of the art deep learning based stereo estimation technique, HD3~\cite{yin2019hierarchical}, as the input to our method. We removed depth values in the depth maps where the reported confidence equals 0.0, and then ran our approach to fill those areas with more accurate depth values. Using our approach we get an improvement of 0.09\% on the foreground objects and 0.04\% overall. HD3 trained and tested on the KITTI dataset outputs complete and accurate depth maps and so we observe small improvement by applying DDP. For sample point of an out-of-distribution model, we used HD3 model trained on the FlyingThings3D dataset~\cite{mayer2016large}. ELAS (which does not use deep learning) does 69 percentage point better. It highlights that our approach with traditional stereo methods can work well under different environment conditions. 
    % \item In the previous setup, HD3 has been trained and tested on the KITTI dataset. However, a depth (or stereo) estimation method should work on any environment. Traditional stereo methods like ELAS~\cite{geiger2010efficient} works without any environment constraints. So, in this experiment, we compare the performance of DDP method applied on ELAS method to that of a deep learning model trained on a out-of-distribution data. For the out-of-distribution model, we used HD3 model trained on the FlyingThings3D dataset~\cite{mayer2016large}. We applied DDP based refinement and completion over depth from ELAS method. In the first setting it generates complete disparity maps without any holes. We use DDP for refinement and completion. We observe a small performance improvement ($\sim$0.1\% reduction in D1-all error). However, compared to the HD3 model trained on the Things dataset both ELAS and ELAS + DDP achieves almost 71 percentage point improvement. It highlights that our approach with traditional stereo methods can work well under different environment conditions. %\VIBHAV{We need to improve this argument. Also why we did not do DDP over HD3 trained on Things output?} 
    % \item The second set of parameters for ELAS generates depth images with holes. In this case, DDP applied over ELAS depth helps to inpaint and complete the ELAS depth. Here we observe a 4\% reduction in D1-all error.
\end{enumerate}

\subsection{Our dataset}
%\vspace{-0.5cm}
\begin{figure}
\centering
  \includegraphics[width=0.95\linewidth]{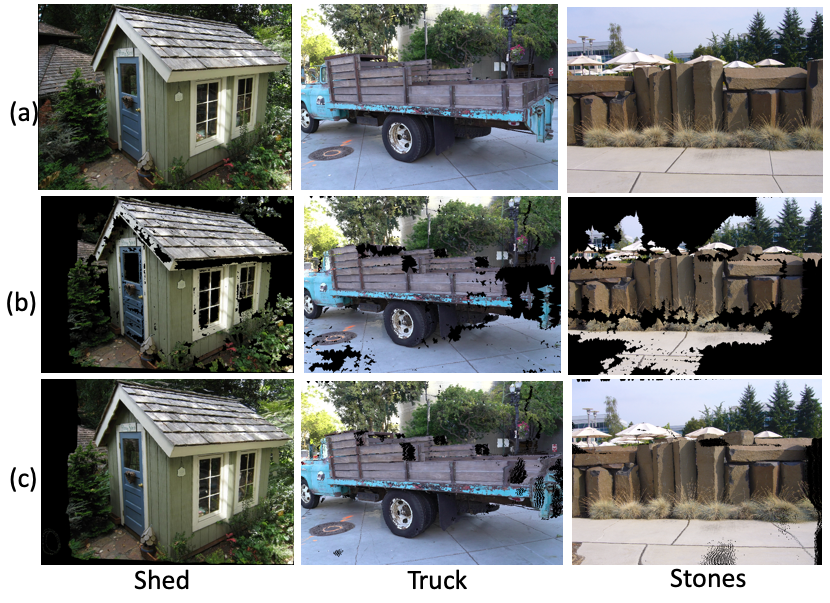}
\caption{(a) Input image from the reference view point (b) Novel view synthesized from neighboring to reference viewpoint using the SGM depth (c) Novel view synthesis from a neighboring to reference viewpoint using DDP depth. The holes that appear in (b) gets filled in (c). (Best viewed in digital. Please zoom in.)}%\neel{figure is a bit small, can you make it bigger. or add zoomed crops?}}
\label{fig:warpedRGB}
%\vspace{-0.5cm}
\end{figure}

\begin{figure}
\centering
\includegraphics[width=0.95\linewidth]{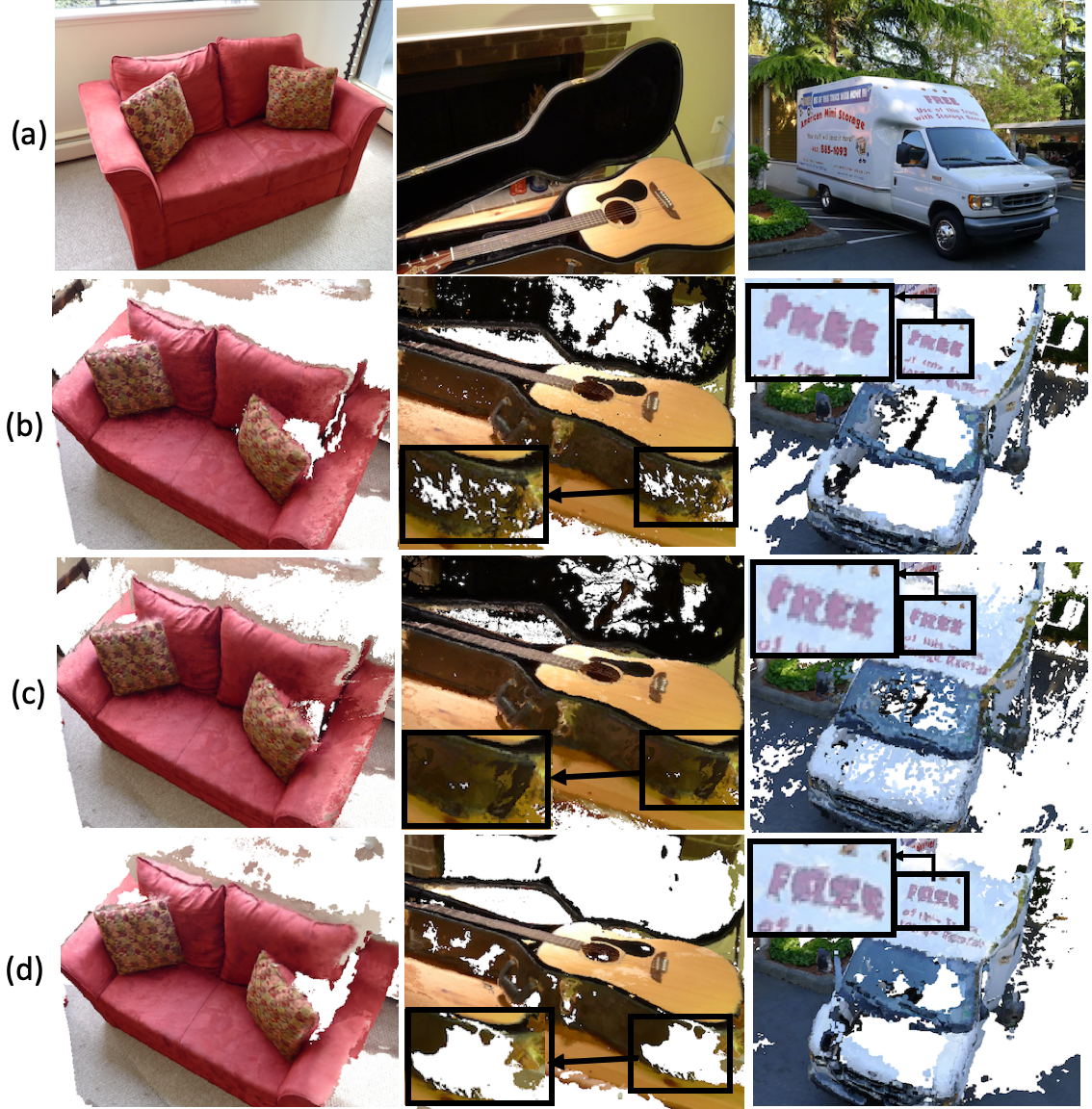}
\caption{(a) Input RGB image  and reconstructed point-cloud from (b) SGM (c) DDP (Ours) and (d) MVSNet depth images for RedCouch, Guitar and Van. Our reconstructions are better and more complete. (Best viewed in digital. Please zoom in.)}
\label{fig:pointcloud}
%\vspace{-0.5cm}
\end{figure}
We used an off-the-shelf consumer camera to capture 5 scenes, both indoor and outdoor. The datasets are monocular image sequences and are named Guitar, Shed, Stones, Red Couch, and Van.

To better understand the quality of the depth images generated by the proposed method, we warp RGB images using the original and the proposed DDP based depth images for Shed and Stones from our video sequences and Truck from TnT dataset. The warped RGB images are shown in the Figure~\ref{fig:warpedRGB}. Holes can be observed in the RGB images warped using original depth maps for example along the roof of the shed, some parts of the truck and along the base of the Stone Wall. However, RGB images warped using our depth images removes large portions of these holes and are far smoother. 

We show reconstructed point clouds on RedCouch, Guitar and Van in Figure~\ref{fig:pointcloud}. %\VIBHAV{How did we record these videos? Are they monocular or stereo videos?} 
The first row is one of the input RGB images used for the reconstruction, second row is the reconstruction from the original SGM output, the third one is from SGM + DDP, and the fourth row is the reconstructed result from MVSNet. The number of views used for these reconstructions are small ($\sim$10). We chose these datasets to show specific ways in which our reconstructions improve on the original SGM and even MVSNet outputs.  As we can see from our results, there are a lot less holes in the reconstructions computed from our depth maps compared to original SGM and MVSNet. For example from the top view of the RedCouch in the first column, we can see the relatively obscure region behind the pillow. The DDP successfully fills up a big portion of this hole. Next the reflective surfaces of the guitar in the second column and the van in the third column also get completely or at least partially filled depending on how big the original hole was.  For example note how the text ``FREE" on the van is more readable in the DDP result. 

%\neel{How come you use 5 datasets in Fig 6 and Fig 7 has 4: 2 from Fig 6 and 2 not used in Fig 6.  I think it would be more consistent if the same datasets were in both.  Maybe too late to change now though.} 
\subsection{NYU v2}

\begin{figure}
\centering
  \includegraphics[width=\linewidth]{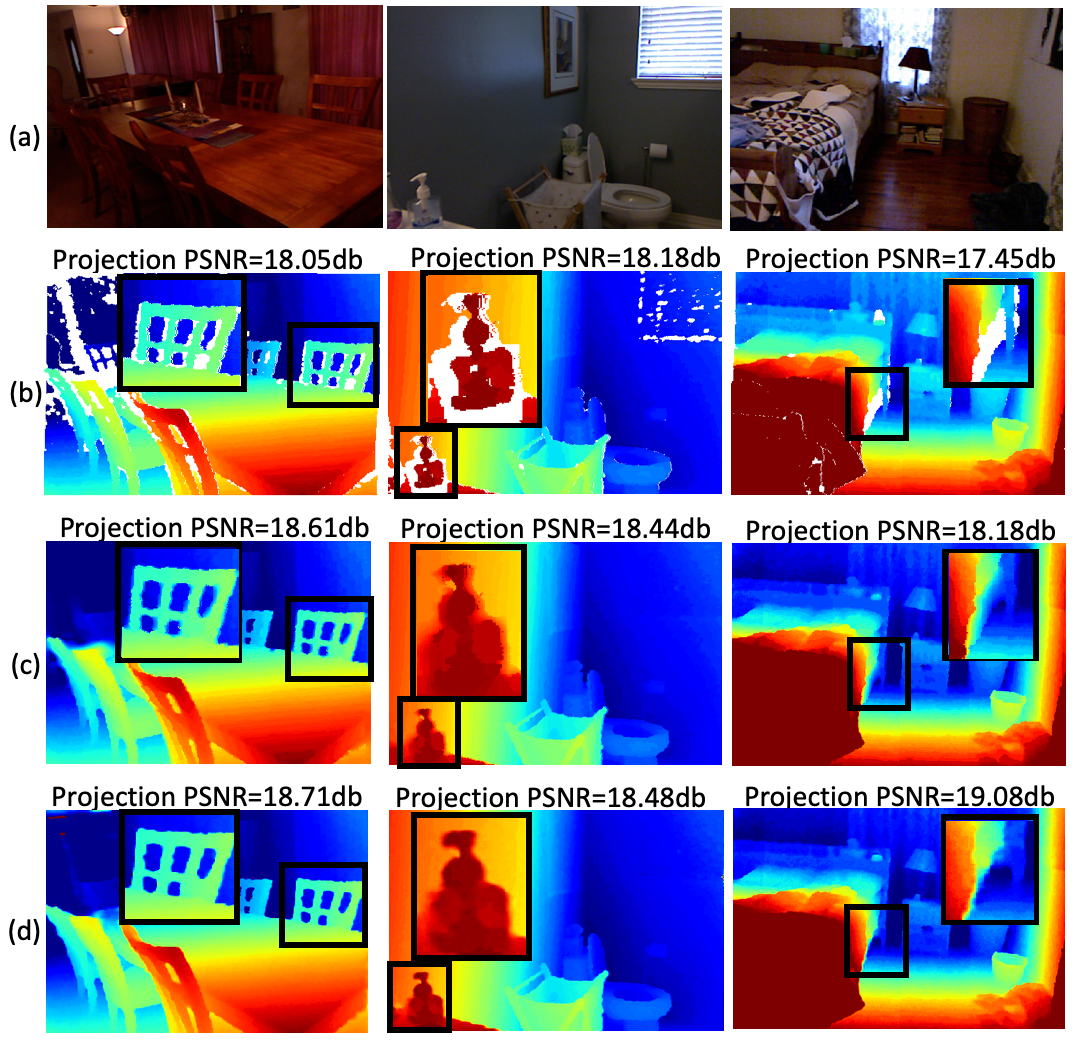}
  \caption{(a) Input RGB (b) Input depth (c) Depth completed using a cross-bilateral filter and (d) Depth completed using DDP. (Best viewed in digital. Please zoom in.)} % \neel{can you but a border around the zoom in, as it is hard to see on some.  Also maybe for the top row, but the crops over the bottom right, so they don't have to go above as they do now?}}
\label{fig:nyu}
\end{figure}

We demonstrate performance of our approach on hole-filling Kinect depth data. For this, we use data from the popular NYU v2 dataset~\cite{eccv_SilbermanHKF12}. Qualitative experiments on the NYU v2 depth images are shown in Figure~\ref{fig:nyu}.
%
%We did some qualitative experiments on NYU depth V2 dataset as shown in Fig. \ref{fig:nyu}. 
%
We cannot directly use the test data from NYU, since we need at least two views as input. So we downloaded 3 video scenes, and used a structure-from-motion (SfM) pipeline \cite{ullman1979interpretation} %\neel{it's it actually Sfm and not SGM?.SGM is for depth maps, SFM will give camera pose} 
to get the extrinsic camera pose information. Using this, we apply DDP to remove holes in the Kinect depth maps given by the dataset. We directly refined the Kinect depth images, and as the originals are incomplete, we cannot use them for performance analysis. Instead we use the depth maps to project one RGB view to another in the video sequences to compute and RGB re-projection error, which we quantify with Peak Signal to noise ratio (PSNR). %(\VIBHAV{What is this?})
It is defined as $PSNR = 20\log_{10}(MAX_{I}/\sqrt{MSE})$, where $MAX_{I}$ is the maximum value of the image and $MSE$ is the mean squared error of the image. As we can see in Figure~\ref{fig:nyu}, DDP fills up holes and improves the PSNR. We also use the cross-bilateral hole filled depth maps included in the NYU v2 dataset as a baseline. We observe consistent qualitative and quantitative improvements using DDP compared to the cross-bilateral method.

\subsection{Middlebury}

Finally we compare to the baseline techniques described in Depthcomp~\cite{atapour2017depthcomp} for Middlebury Dataset scenes \cite{hirschmuller2007evaluation,scharstein2007learning} in Table~\ref{table:completion_baseline}. We took the input images and disparity maps from Depthcomp~\cite{atapour2017depthcomp} with the holes they created artificially in the disparity maps and compared to the baseline techniques they mention. The metric is Root Mean Square Error (RMSE) which is $||D_\text{out} - D_\text{GT}||_{2}$ where $D_\text{out}$ is the output disparity and $D_\text{GT}$ is the ground-truth disparity. Our average result is better than all of the other techniques in Table~\ref{table:completion_baseline} for depth completion.

\begin{table}[t]
\centering
\caption{Comparison of our method with techniques mentioned in~\cite{atapour2017depthcomp} on Middlebury dataset using RMSE. Lower is better.}
\begin{tabular}{@{}lcccc@{}}
\toprule
Method & Plastic & Baby & Bowling & Average\\
\midrule
SSI \cite{herrera2013depth} & 1.7573 & 2.9638 & 6.4936 & {\it 3.74}\\
Linear Inter & 1.3432 & 1.3473 & 1.4503 & {\it 1.38}\\
Cubic Inter & 1.2661 & 1.3384 & 1.4460 & {\it 1.35} \\
FMM \cite{telea2004image} & 0.9580 & 0.8349 & 1.2422 & {\it 1.01}\\
GIF \cite{liu2012guided} & 0.7947 & 0.6008 & 0.9436 & {\it 0.78}\\
FBF \cite{atapour2016back} & 0.8643 & 0.6238 & 0.5918 & {\it 0.69}\\
EBI \cite{arias2011variational} & 0.6952 & 0.6755 & 0.4857 & {\it 0.62}\\
Depthcomp \cite{atapour2017depthcomp} & 0.6618 & 0.3697 & 0.4292 & {\it 0.49}\\
{\bf DDP (Ours)} & {\bf 0.4951} & {\bf 0.3232} & {\bf 0.5743 } & {\bf 0.46}\\
\bottomrule
\end{tabular}

\label{table:completion_baseline}
\end{table}

\section{Conclusions}

We have presented an approach to reconstruct depth maps from incomplete ones. %depth map and a color image taken from the same viewpoint. 
We leverage the recently proposed idea of utilizing a neural network as a prior for natural color images, and introduced three new loss terms for %reconstruct clean and 
depth map completion. Extensive qualitative and quantitative experiments on sequences from the Tanks and Temples, KITTI, NYU v2, Middlebury, and our own dataset demonstrated that the depth maps generated by our method were more accurate. %One 
An important future extension is improving the speed of the method, where an efficient version of the presented approach could be used for a real-time depth enhancement and 3D reconstruction pipeline. Further, the presented method could benefit from deeper understanding of the convergence properties of training deep image priors. 

\begin{figure*}
\centering

 \includegraphics[width=0.88\linewidth]{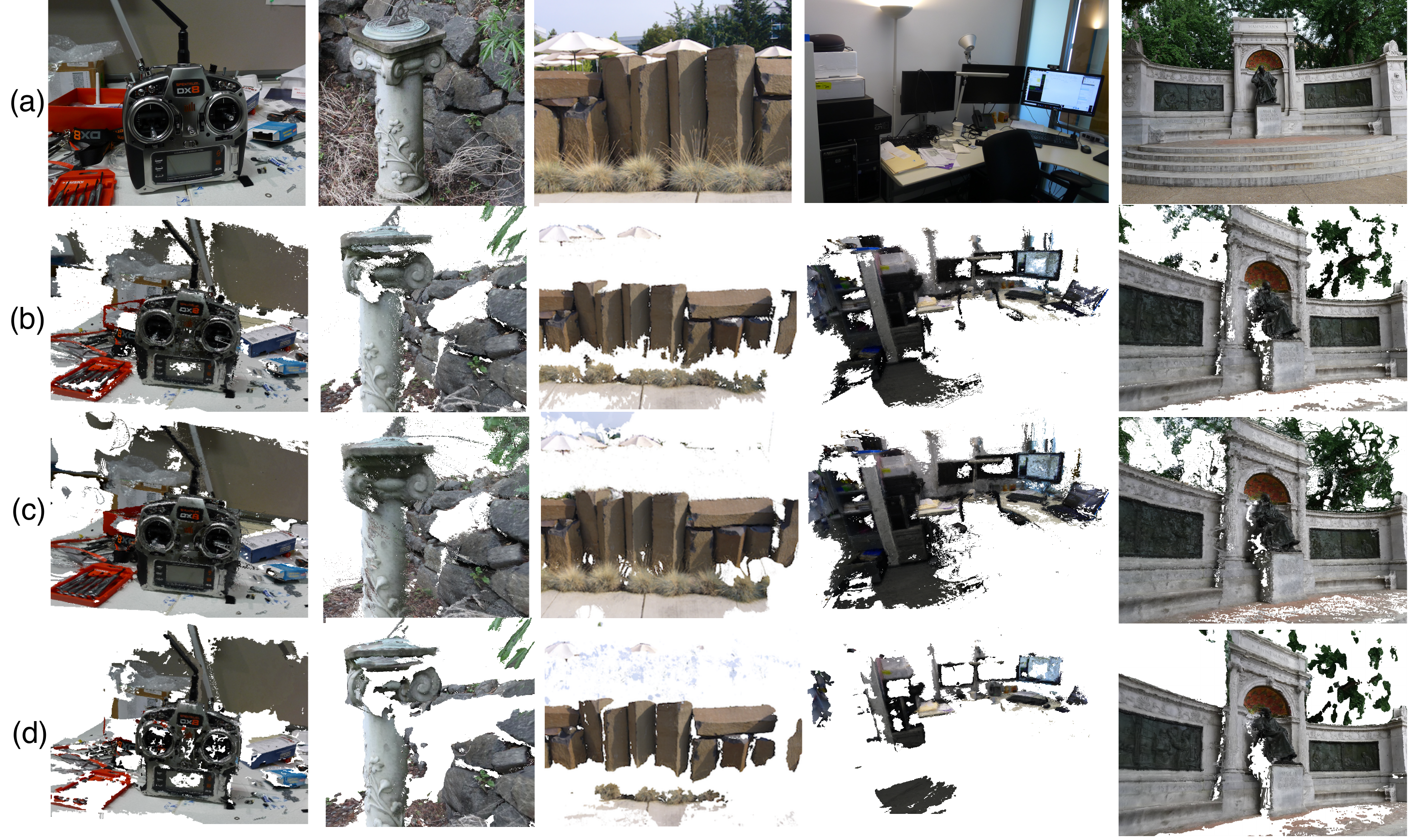}

\caption{(a) Input RGB image  and reconstructed pointcloud from (b) SGM (c) DDP (Ours) and (d) MVSNet depth images for Desk, Sundial, Stones, Office and Statue. Our reconstructions are more complete.}
\label{Fig:pointcloud2}
\end{figure*}

%\vfill\clearpage

\begin{center}
\title{\bf \Large Supplementary}
\maketitle
\end{center}

\section{Reconstruction Pipeline}
\label{section:pipeline}

To illustrate the impact of the presented approach in the method section of the main paper, we show improvement in quality of reconstruction from a set of images. We leverage the improved depth images generated by our approach within standard multi-view reconstruction pipeline \cite{galliani2015massively}. Some of the improved models generated by our method is shown in the Figure \ref{Fig:pointcloud2} as well as in the main paper.

%\Vibhav{@Sudipta, @Pallabi, could you please complete the pipeline stage?}
High quality depth images are first generated using the presented deep depth prior method on SGM~\cite{SGM} depths. These depth images are then fused using Fusibile approach~\cite{galliani2015massively} to reconstruct the final 3d point cloud. Fusibile has certain hyperparameters that determine the precision and recall values for the resultant point cloud. These parameters include the disparity threshold and the number of consistent views. The disparity threshold determines the threshold of difference in disparity  that is allowed for two points from two different depth maps to be merged. The number of consistent images shows the number of images in which a point has to have consistent depth values for it to appear in the merged point cloud. For different datasets we observed that different values for these hyperparameters gave the best results for SGM and our results. We report these hyperparameter values in Table 3 in the main paper as well. 

%We also use per-pixel depth uncertainty estimates from the SGM method to remove depth points with uncertainty values above certain threshold (0.2 or 0.5) during reconstruction in some scenes.

\section{Hyperparameters}

In equation 2 in main paper
for TnT~\cite{knapitsch2017tanks} and our dataset:
$\gamma_{1}$ = 0.96
$\gamma_{2}$ = 0.02; 
for all other dataset:
$\gamma_{1}$ = 0.98
$\gamma_{2}$ = 0.01.
In equation 3 in main paper:
$\lambda_{z}$ = 0.8. 
In equation 4 in main paper:
$\lambda_{I}$ = 0.5. 
In equation 5 in main paper:
$\lambda_{w}$ = 0.5. 
Hyper-parameters of the loss function are set through searching on a small subset (10 images randomly chosen) from each of Ignatius and Barn sequence in TnT and the validation set from KITTI~\cite{geiger2013vision,Menze2015CVPR}. 
We use $N$=2 nearest neighbors while calculating warping based loss for multi-view datasets. For DDP we use a maximum and a minimum depth value to clip the depth maps for 7 different scenes in TnT dataset. We also add a constant value to all depths as mentioned in the main paper. These values are in Table~\ref{table:min_max_dep}. For TnT and our dataset, the initial learning rate is set to 0.00005, that is reduced by a factor of 0.01 after 12000 and 15000 epochs. The model is trained for 16000 epochs. For KITTI and NYU~\cite{eccv_SilbermanHKF12} datasets, the model is trained for 10000 epochs with initial learning rate of 0.00005. %For Middlebury dataset~\cite{hirschmuller2007evaluation,scharstein2007learning} we run the training for 16000 epochs at initial learning rate of 0.0001.

\begin{table}
\centering
\caption{Minimum and maximum depth clipping values and the constant depth value added per scene before doing DDP for 7 scenes in TnT dataset}
\label{table:min_max_dep}
\resizebox{0.8\linewidth}{!}{
    \begin{tabular}{@{}lcccc@{}}
    \toprule
    & min depth & max depth & constant \\
    \midrule
    Ignatius & 2.0 & 7.5 & 0.0 \\
    Barn & 2.0 & 16.5 & 2.0 \\
    Caterpillar & 2.0 & 7.5 & 0.0 \\
    Meetingroom & 0.2 & 25.0 & 4.0 \\
    Truck & 0.5 & 10.0 & 2.0 \\
    Courtroom & 0.2 & 46.0 & 4.0 \\
    Church & 0.2 & 16.0 & 4.0\\
    \bottomrule
    \end{tabular}}
\end{table}

\section{SSIM Index}

Structural Similarity index (SSIM) \cite{ssim2004tpami} is defined by the the equation~\ref{eq:s1},
\vspace{-2mm}
\begin{equation}
\begin{aligned}
SSIM(x,y) = \frac{(2\mu_{x}\mu_{y}+c_{1})(2\sigma_{xy}+c_{2})}{(\mu_{x}^{2}+\mu_{y}^{2}+c_{1})(\sigma_{x}^{2}+\sigma_{y}^{2}+c_{2})}
\end{aligned}
\label{eq:s1}
\end{equation}
where,
$\mu_{x}$ and $\mu_{y}$ are the averages of x and y,
$\sigma_{x}^{2}$ and $\sigma_{y}^{2}$ are the variances of x and y,
$\sigma_{xy}$ is the covariance of x and y,
$c_{1}= (k_{1}L)^{2}$ and $c_{2}=(k_{2}L)^{2}$ where $L$ = dynamic range of pixel values and $k_{1}$ and $k_{2}$ are constants.

\section{Qualitative Results}

Along with the reconstructed models in the main paper we add 5 more results here in Figure~\ref{Fig:pointcloud2}. These 5 scenes are Desk, Sundial, Stones, Office and Statue respectively which we collect ourselves using the same technique as described in Section 4.3 in the main paper. We compare SGM depth based reconstruction, DDP (ours) depth based reconstruction and MVSNet~\cite{yao2018mvsnet} depth based reconstruction. The central portion of the radio on the desk gets filled in our reconstruction as well as the red tool box. The hidden portion of the sundial and the base of stones have holes in the SGM based reconstruction that gets partially filled in ours. The reflective computer screens and the holes on reflective desk in the Office scene gets partially filled. Finally we can see more trees and smaller holes in the Statue scene. 
We observe some holes in our reconstructions, in spite of the depth maps being complete. This is because the Fusibile parameters described in Section~\ref{section:pipeline} remove inconsistent depths from multiple views.

{\small
\bibliographystyle{ieee}
\bibliography{egbib}

\begin{thebibliography}{10}\itemsep=-1pt

\bibitem{aanaes2016large}
H.~Aan{\ae}s, R.~R. Jensen, G.~Vogiatzis, E.~Tola, and A.~B. Dahl.
\newblock Large-scale data for multiple-view stereopsis.
\newblock {\em International Journal of Computer Vision}, pages 1--16, 2016.

\bibitem{arias2011variational}
P.~Arias, G.~Facciolo, V.~Caselles, and G.~Sapiro.
\newblock A variational framework for exemplar-based image inpainting.
\newblock {\em International journal of computer vision}, 93(3):319--347, 2011.

\bibitem{atapour2017depthcomp}
A.~Atapour-Abarghouei and T.~P. Breckon.
\newblock Depthcomp: real-time depth image completion based on prior semantic
  scene segmentation.
\newblock 2017.

\bibitem{atapour2016back}
A.~Atapour-Abarghouei, G.~P. de~La~Garanderie, and T.~P. Breckon.
\newblock Back to butterworth-a fourier basis for 3d surface relief hole
  filling within rgb-d imagery.
\newblock In {\em 2016 23rd International Conference on Pattern Recognition
  (ICPR)}, pages 2813--2818. IEEE, 2016.

\bibitem{sgmGPU}
C.~Banz, H.~Blume, and P.~Pirsch.
\newblock Real-time semi-global matching disparity estimation on the gpu.
\newblock In {\em 2011 IEEE International Conference on Computer Vision
  Workshops (ICCV Workshops)}, pages 514--521. IEEE, 2011.

\bibitem{barron2016fast}
J.~T. Barron and B.~Poole.
\newblock The fast bilateral solver.
\newblock In {\em European Conference on Computer Vision}, pages 617--632.
  Springer, 2016.

\bibitem{bleyer2011patchmatch}
M.~Bleyer, C.~Rhemann, and C.~Rother.
\newblock Patchmatch stereo-stereo matching with slanted support windows.
\newblock In {\em Bmvc}, volume~11, pages 1--11, 2011.

\bibitem{bleyer2011object}
M.~Bleyer, C.~Rother, P.~Kohli, D.~Scharstein, and S.~Sinha.
\newblock Object stereo—joint stereo matching and object segmentation.
\newblock In {\em CVPR 2011}, pages 3081--3088. IEEE, 2011.

\bibitem{chang2018pyramid}
J.-R. Chang and Y.-S. Chen.
\newblock Pyramid stereo matching network.
\newblock In {\em Proceedings of the IEEE Conference on Computer Vision and
  Pattern Recognition}, pages 5410--5418, 2018.

\bibitem{chen2015deep}
Z.~Chen, X.~Sun, L.~Wang, Y.~Yu, and C.~Huang.
\newblock A deep visual correspondence embedding model for stereo matching
  costs.
\newblock In {\em Proceedings of the IEEE International Conference on Computer
  Vision}, pages 972--980, 2015.

\bibitem{cheng2019bayesian}
Z.~Cheng, M.~Gadelha, S.~Maji, and D.~Sheldon.
\newblock A bayesian perspective on the deep image prior.
\newblock In {\em CVPR}, pages 5443--5451, 2019.

\bibitem{dai2017scannet}
A.~Dai, A.~X. Chang, M.~Savva, M.~Halber, T.~Funkhouser, and M.~Nie{\ss}ner.
\newblock Scannet: Richly-annotated 3d reconstructions of indoor scenes.
\newblock In {\em Proceedings of the IEEE Conference on Computer Vision and
  Pattern Recognition}, pages 5828--5839, 2017.

\bibitem{furukawa2010towards}
Y.~Furukawa, B.~Curless, S.~M. Seitz, and R.~Szeliski.
\newblock Towards internet-scale multi-view stereo.
\newblock In {\em 2010 IEEE computer society conference on computer vision and
  pattern recognition}, pages 1434--1441. IEEE, 2010.

\bibitem{fusible}
S.~{Galliani}, K.~{Lasinger}, and K.~{Schindler}.
\newblock Massively parallel multiview stereopsis by surface normal diffusion.
\newblock In {\em 2015 IEEE International Conference on Computer Vision
  (ICCV)}, pages 873--881, Dec 2015.

\bibitem{galliani2015massively}
S.~Galliani, K.~Lasinger, and K.~Schindler.
\newblock Massively parallel multiview stereopsis by surface normal diffusion.
\newblock In {\em Proceedings of the IEEE International Conference on Computer
  Vision}, pages 873--881, 2015.

\bibitem{gandelsman2018double}
Y.~Gandelsman, A.~Shocher, and M.~Irani.
\newblock ``{D}ouble-{DIP}": Unsupervised image decomposition via coupled
  deep-image-priors.
\newblock In {\em CVPR}, 2019.

\bibitem{gehrig2009real}
S.~K. Gehrig, F.~Eberli, and T.~Meyer.
\newblock A real-time low-power stereo vision engine using semi-global
  matching.
\newblock In {\em International Conference on Computer Vision Systems}, pages
  134--143. Springer, 2009.

\bibitem{geiger2013vision}
A.~Geiger, P.~Lenz, C.~Stiller, and R.~Urtasun.
\newblock Vision meets robotics: The kitti dataset.
\newblock {\em The International Journal of Robotics Research},
  32(11):1231--1237, 2013.

\bibitem{geiger2010efficient}
A.~Geiger, M.~Roser, and R.~Urtasun.
\newblock Efficient large-scale stereo matching.
\newblock In {\em Asian conference on computer vision}, pages 25--38. Springer,
  2010.

\bibitem{goesele2007multi}
M.~Goesele, N.~Snavely, B.~Curless, H.~Hoppe, and S.~M. Seitz.
\newblock Multi-view stereo for community photo collections.
\newblock In {\em 2007 IEEE 11th International Conference on Computer Vision},
  pages 1--8. IEEE, 2007.

\bibitem{mvSGM}
N.~Haala and M.~Rothermel.
\newblock Dense multi-stereo matching for high quality digital elevation
  models.
\newblock {\em Photogrammetrie-Fernerkundung-Geoinformation}, 2012(4):331--343,
  2012.

\bibitem{heise2013pm}
P.~Heise, S.~Klose, B.~Jensen, and A.~Knoll.
\newblock Pm-huber: Patchmatch with huber regularization for stereo matching.
\newblock In {\em Proceedings of the IEEE International Conference on Computer
  Vision}, pages 2360--2367, 2013.

\bibitem{herrera2013depth}
D.~Herrera, J.~Kannala, J.~Heikkil{\"a}, et~al.
\newblock Depth map inpainting under a second-order smoothness prior.
\newblock In {\em Scandinavian Conference on Image Analysis}, pages 555--566.
  Springer, 2013.

\bibitem{SGM}
H.~Hirschmuller.
\newblock Stereo processing by semiglobal matching and mutual information.
\newblock {\em IEEE Transactions on pattern analysis and machine intelligence},
  30(2):328--341, 2007.

\bibitem{hirschmuller2007evaluation}
H.~Hirschmuller and D.~Scharstein.
\newblock Evaluation of cost functions for stereo matching.
\newblock In {\em 2007 IEEE Conference on Computer Vision and Pattern
  Recognition}, pages 1--8. IEEE, 2007.

\bibitem{huang2018deepmvs}
P.-H. Huang, K.~Matzen, J.~Kopf, N.~Ahuja, and J.-B. Huang.
\newblock Deepmvs: Learning multi-view stereopsis.
\newblock In {\em Proceedings of the IEEE Conference on Computer Vision and
  Pattern Recognition}, pages 2821--2830, 2018.

\bibitem{ji2017surfacenet}
M.~Ji, J.~Gall, H.~Zheng, Y.~Liu, and L.~Fang.
\newblock Surfacenet: An end-to-end 3d neural network for multiview stereopsis.
\newblock In {\em Proceedings of the IEEE Conference on Computer Vision and
  Pattern Recognition}, pages 2307--2315, 2017.

\bibitem{kendall2017end}
A.~Kendall, H.~Martirosyan, S.~Dasgupta, P.~Henry, R.~Kennedy, A.~Bachrach, and
  A.~Bry.
\newblock End-to-end learning of geometry and context for deep stereo
  regression.
\newblock In {\em Proceedings of the IEEE International Conference on Computer
  Vision}, pages 66--75, 2017.

\bibitem{knapitsch2017tanks}
A.~Knapitsch, J.~Park, Q.-Y. Zhou, and V.~Koltun.
\newblock Tanks and temples: Benchmarking large-scale scene reconstruction.
\newblock {\em ACM Transactions on Graphics (ToG)}, 36(4):78, 2017.

\bibitem{knobelreiter2019learned}
P.~Kn{\"o}belreiter and T.~Pock.
\newblock Learned collaborative stereo refinement.
\newblock In {\em German Conference on Pattern Recognition}, pages 3--17.
  Springer, 2019.

\bibitem{knobelreiter2017end}
P.~Knobelreiter, C.~Reinbacher, A.~Shekhovtsov, and T.~Pock.
\newblock End-to-end training of hybrid cnn-crf models for stereo.
\newblock In {\em Proceedings of the IEEE Conference on Computer Vision and
  Pattern Recognition}, pages 2339--2348, 2017.

\bibitem{liu2012guided}
J.~Liu, X.~Gong, and J.~Liu.
\newblock Guided inpainting and filtering for kinect depth maps.
\newblock In {\em Proceedings of the 21st International Conference on Pattern
  Recognition (ICPR2012)}, pages 2055--2058. IEEE, 2012.

\bibitem{lu2013patch}
J.~Lu, H.~Yang, D.~Min, and M.~N. Do.
\newblock Patch match filter: Efficient edge-aware filtering meets randomized
  search for fast correspondence field estimation.
\newblock In {\em Proceedings of the IEEE conference on computer vision and
  pattern recognition}, pages 1854--1861, 2013.

\bibitem{luo2016efficient}
W.~Luo, A.~G. Schwing, and R.~Urtasun.
\newblock Efficient deep learning for stereo matching.
\newblock In {\em Proceedings of the IEEE Conference on Computer Vision and
  Pattern Recognition}, pages 5695--5703, 2016.

\bibitem{mayer2016large}
N.~Mayer, E.~Ilg, P.~Hausser, P.~Fischer, D.~Cremers, A.~Dosovitskiy, and
  T.~Brox.
\newblock A large dataset to train convolutional networks for disparity,
  optical flow, and scene flow estimation.
\newblock In {\em Proceedings of the IEEE Conference on Computer Vision and
  Pattern Recognition}, pages 4040--4048, 2016.

\bibitem{Menze2015CVPR}
M.~Menze and A.~Geiger.
\newblock Object scene flow for autonomous vehicles.
\newblock In {\em Conference on Computer Vision and Pattern Recognition
  (CVPR)}, 2015.

\bibitem{pang2017cascade}
J.~Pang, W.~Sun, J.~S. Ren, C.~Yang, and Q.~Yan.
\newblock Cascade residual learning: A two-stage convolutional neural network
  for stereo matching.
\newblock In {\em Proceedings of the IEEE International Conference on Computer
  Vision}, pages 887--895, 2017.

\bibitem{ren2017unsupervised}
Z.~Ren, J.~Yan, B.~Ni, B.~Liu, X.~Yang, and H.~Zha.
\newblock Unsupervised deep learning for optical flow estimation.
\newblock In {\em Thirty-First AAAI Conference on Artificial Intelligence},
  2017.

\bibitem{ronneberger2015u}
O.~Ronneberger, P.~Fischer, and T.~Brox.
\newblock U-net: Convolutional networks for biomedical image segmentation.
\newblock In {\em International Conference on Medical image computing and
  computer-assisted intervention}, pages 234--241. Springer, 2015.

\bibitem{scharstein2014high}
D.~Scharstein, H.~Hirschm{\"u}ller, Y.~Kitajima, G.~Krathwohl,
  N.~Ne{\v{s}}i{\'c}, X.~Wang, and P.~Westling.
\newblock High-resolution stereo datasets with subpixel-accurate ground truth.
\newblock In {\em German conference on pattern recognition}, pages 31--42.
  Springer, 2014.

\bibitem{scharstein2007learning}
D.~Scharstein and C.~Pal.
\newblock Learning conditional random fields for stereo.
\newblock In {\em 2007 IEEE Conference on Computer Vision and Pattern
  Recognition}, pages 1--8. IEEE, 2007.

\bibitem{middlebury}
D.~Scharstein and R.~Szeliski.
\newblock A taxonomy and evaluation of dense two-frame stereo correspondence
  algorithms.
\newblock {\em Int. J. Comput. Vision}, 47(1–3):7–42, Apr. 2002.

\bibitem{Scharstein2017}
D.~Scharstein, T.~Taniai, and S.~N. Sinha.
\newblock Semi-global stereo matching with surface orientation priors.
\newblock {\em 2017 International Conference on 3D Vision (3DV)}, pages
  215--224, 2017.

\bibitem{COLMAPb}
J.~L. Sch\"{o}nberger, E.~Zheng, M.~Pollefeys, and J.-M. Frahm.
\newblock Pixelwise view selection for unstructured multi-view stereo.
\newblock In {\em European Conference on Computer Vision (ECCV)}, 2016.

\bibitem{schops2017multi}
T.~Schops, J.~L. Schonberger, S.~Galliani, T.~Sattler, K.~Schindler,
  M.~Pollefeys, and A.~Geiger.
\newblock A multi-view stereo benchmark with high-resolution images and
  multi-camera videos.
\newblock In {\em Proceedings of the IEEE Conference on Computer Vision and
  Pattern Recognition}, pages 3260--3269, 2017.

\bibitem{seki2017sgm}
A.~Seki and M.~Pollefeys.
\newblock Sgm-nets: Semi-global matching with neural networks.
\newblock In {\em Proceedings of the IEEE Conference on Computer Vision and
  Pattern Recognition}, pages 231--240, 2017.

\bibitem{eccv_SilbermanHKF12}
N.~Silberman, D.~Hoiem, P.~Kohli, and R.~Fergus.
\newblock Indoor segmentation and support inference from {RGBD} images.
\newblock In {\em Computer Vision - {ECCV} 2012 - 12th European Conference on
  Computer Vision, Florence, Italy, October 7-13, 2012, Proceedings, Part {V}},
  volume 7576 of {\em Lecture Notes in Computer Science}, pages 746--760.
  Springer, 2012.

\bibitem{sinha2014efficient}
S.~N. Sinha, D.~Scharstein, and R.~Szeliski.
\newblock Efficient high-resolution stereo matching using local plane sweeps.
\newblock In {\em Proceedings of the IEEE Conference on Computer Vision and
  Pattern Recognition}, pages 1582--1589, 2014.

\bibitem{song2015sun}
S.~Song, S.~P. Lichtenberg, and J.~Xiao.
\newblock Sun rgb-d: A rgb-d scene understanding benchmark suite.
\newblock In {\em Proceedings of the IEEE conference on computer vision and
  pattern recognition}, pages 567--576, 2015.

\bibitem{taniai2018neural}
T.~Taniai and T.~Maehara.
\newblock Neural inverse rendering for general reflectance photometric stereo.
\newblock In {\em ICML}, 2018.

\bibitem{taniai2014graph}
T.~Taniai, Y.~Matsushita, and T.~Naemura.
\newblock Graph cut based continuous stereo matching using locally shared
  labels.
\newblock In {\em Proceedings of the IEEE Conference on Computer Vision and
  Pattern Recognition}, pages 1613--1620, 2014.

\bibitem{taniai2017continuous}
T.~Taniai, Y.~Matsushita, Y.~Sato, and T.~Naemura.
\newblock Continuous 3d label stereo matching using local expansion moves.
\newblock {\em IEEE transactions on pattern analysis and machine intelligence},
  40(11):2725--2739, 2017.

\bibitem{telea2004image}
A.~Telea.
\newblock An image inpainting technique based on the fast marching method.
\newblock {\em Journal of graphics tools}, 9(1):23--34, 2004.

\bibitem{tonioni2017unsupervised}
A.~Tonioni, M.~Poggi, S.~Mattoccia, and L.~Di~Stefano.
\newblock Unsupervised adaptation for deep stereo.
\newblock In {\em Proceedings of the IEEE International Conference on Computer
  Vision}, pages 1605--1613, 2017.

\bibitem{ullman1979interpretation}
S.~Ullman.
\newblock The interpretation of structure from motion.
\newblock {\em Proceedings of the Royal Society of London. Series B. Biological
  Sciences}, 203(1153):405--426, 1979.

\bibitem{ulyanov2018deep}
D.~Ulyanov, A.~Vedaldi, and V.~Lempitsky.
\newblock Deep image prior.
\newblock In {\em CVPR}, pages 9446--9454, 2018.

\bibitem{voynov2019perceptual}
O.~Voynov, A.~Artemov, V.~Egiazarian, A.~Notchenko, G.~Bobrovskikh, E.~Burnaev,
  and D.~Zorin.
\newblock Perceptual deep depth super-resolution.
\newblock In {\em Proceedings of the IEEE International Conference on Computer
  Vision}, pages 5653--5663, 2019.

\bibitem{ssim2004tpami}
Z.~Wang, A.~Bovik, H.~Sheikh, and E.~Simoncelli.
\newblock Image quality assessment: From error visibility to structural
  similarity.
\newblock In {\em IEEE transactions on image processing}, page 600–612. IEEE,
  2004.

\bibitem{williams2019deep}
F.~Williams, T.~Schneider, C.~Silva, D.~Zorin, J.~Bruna, and D.~Panozzo.
\newblock Deep geometric prior for surface reconstruction.
\newblock In {\em CVPR}, pages 10130--10139, 2019.

\bibitem{woodford2009global}
O.~Woodford, P.~Torr, I.~Reid, and A.~Fitzgibbon.
\newblock Global stereo reconstruction under second-order smoothness priors.
\newblock {\em IEEE transactions on pattern analysis and machine intelligence},
  31(12):2115--2128, 2009.

\bibitem{xue2019mvscrf}
Y.~Xue, J.~Chen, W.~Wan, Y.~Huang, C.~Yu, T.~Li, and J.~Bao.
\newblock Mvscrf: Learning multi-view stereo with conditional random fields.
\newblock In {\em Proceedings of the IEEE International Conference on Computer
  Vision}, pages 4312--4321, 2019.

\bibitem{yao2018mvsnet}
Y.~Yao, Z.~Luo, S.~Li, T.~Fang, and L.~Quan.
\newblock Mvsnet: Depth inference for unstructured multi-view stereo.
\newblock In {\em Proceedings of the European Conference on Computer Vision
  (ECCV)}, pages 767--783, 2018.

\bibitem{yin2019hierarchical}
Z.~Yin, T.~Darrell, and F.~Yu.
\newblock Hierarchical discrete distribution decomposition for match density
  estimation.
\newblock In {\em Proceedings of the IEEE Conference on Computer Vision and
  Pattern Recognition}, pages 6044--6053, 2019.

\bibitem{vzbontar2016stereo}
J.~{\v{Z}}bontar and Y.~LeCun.
\newblock Stereo matching by training a convolutional neural network to compare
  image patches.
\newblock {\em The journal of machine learning research}, 17(1):2287--2318,
  2016.

\bibitem{zhang2018deep}
Y.~Zhang and T.~Funkhouser.
\newblock Deep depth completion of a single rgb-d image.
\newblock In {\em Proceedings of the IEEE Conference on Computer Vision and
  Pattern Recognition}, pages 175--185, 2018.

\bibitem{zhou2017}
C.~{Zhou}, H.~{Zhang}, X.~{Shen}, and J.~{Jia}.
\newblock Unsupervised learning of stereo matching.
\newblock In {\em 2017 IEEE International Conference on Computer Vision
  (ICCV)}, pages 1576--1584, Oct 2017.

\end{thebibliography}
}

\end{document}